\documentclass[letterpaper]{article}

\usepackage[round]{natbib}

\usepackage{graphicx}
\usepackage{subfigure}
\usepackage{amsfonts}
\usepackage{amsmath,amsgen,amstext,amssymb}
\usepackage{amsthm}
\usepackage{etoolbox}
\usepackage{algorithm}
\usepackage{algorithmic}
\usepackage{enumitem}
\usepackage{arydshln}
\usepackage{moresize}

\newtoggle{isArxiv}
\togglefalse{isArxiv} 

\newcommand{\citeN}[1]{{\citet{#1}}}

\newcommand{\cL}{{\cal L}}
\newcommand{\cM}{{\cal M}}
\newcommand{\cN}{{\cal N}}
\newcommand{\cP}{{\cal P}}

\newtheorem{theorem}{Theorem}
\newtheorem{lemma}[theorem]{Lemma}

\newtheorem{assumption}{Assumption}

\newtheorem{proposition}[theorem]{Proposition}

\newcommand{\ProofOf}[1] {{\noindent \bf Proof of~#1}}

\newcommand{\refeq}[1] {(\ref{#1})}

\newcommand{\refthm}[1]   {{Theorem~\ref{#1}}}

\newcommand{\refprop}[1]  {{Proposition~\ref{#1}}}
\newcommand{\reflem}[1]  {{Lemma~\ref{#1}}}

\newcommand{\refasm}[1]   {{Assumption~\ref{#1}}}
\newcommand{\refalg}[1]  {{Algorithm~\ref{#1}}}
\newcommand{\reffig}[1]  {{Figure~\ref{#1}}}

\newcommand{\E}{\mathbb{E}}
\newcommand{\Prb}{\mathbb{P}}
\newcommand{\tndi}{\rightarrow \infty}
\newcommand{\tndni}{\rightarrow -\infty}
\newcommand{\tndo}{\rightarrow 0}
\newcommand{\var}{\mbox{Var}}

\newcommand{\DefAs}{:=}

\newcommand{\rob}{{\text{rob}}}

\newcommand{\setX}{\mathbb{X}}
\newcommand{\Reals}{\mathbb{R}}

\begin{document}

%

%

\title{Unbiased Gradient Estimation for Distributionally Robust Learning}

\author{ Soumyadip Ghosh and Mark S.~Squillante\\
Mathematical Sciences, IBM Research AI\\
Thomas J.~Watson Research Center\\
 Yorktown Heights, NY 20198, USA.\\
\texttt{\{ghoshs,mss\}@us.ibm.com}
}

\maketitle

\begin{abstract}
Seeking to improve model generalization, we consider a new approach based on distributionally robust learning (DRL)
that applies stochastic gradient descent to the outer minimization problem.
Our algorithm efficiently estimates the gradient of the inner maximization problem through multi-level Monte Carlo randomization.
Leveraging theoretical results that shed light on why standard gradient estimators fail,
we establish the optimal parameterization of the gradient estimators of our approach that
balances a fundamental tradeoff between computation time and statistical variance.
Numerical experiments demonstrate that our DRL approach yields significant benefits over previous work.
\end{abstract}

\section{Introduction}
The formulation of distributionally robust optimization problems for machine learning (DRL)
has the potential to significantly improve model generalization by reducing the probability
of poor performance over samples not encountered in training.
The minimum-maximum nature of these formulations, however, creates fundamental difficulties
for standard algorithms such as stochastic gradient descent (SGD).
To address these difficulties, we consider in this paper a new SGD algorithm where gradient descent
is applied to the outer minimization problem.
Our main contributions include efficiently estimating the gradient of the inner maximization
problem by exploiting a randomization technique called multi-level Monte Carlo, introduced by~\citeN{giles08}.
We present theoretical results that establish why standard gradient estimators fail and
that determine the range of parameter values for the gradient estimator of our proposed DRL approach which balances
a fundamental tradeoff between stochastic error and computation time.
A set of numerical experiments are also presented demonstrating that our DRL approach yields significant computational savings over previous work,
while further illustrating the efficacy of our DRL approach for better model generalization.

We now provide our formal general setting for the DRL problem, which is consistent with previous recent work in the literature (see, e.g.,~\citeN{nd16,nd17,gsw19}).
Let $\setX$ denote a sample space, $P$ denote a probability distribution on $\setX$, and $\Theta \subseteq \Reals^d$ denote a parameter space.
We define $L_P(\theta) \DefAs \E_P[l(\theta,\xi)]$ to be the expectation, with respect to $P$, of a loss function
$l : \Theta \times \setX \rightarrow \Reals$ which denotes the estimation error for a learning model with parameters
$\theta \in \Theta$ over data $\xi \in \setX$.
We further define the robust loss to be the worst-case expected loss function
$R(\theta) \DefAs \E_{P^*(\theta)}[ l(\theta,\xi) ] = \sup_{P\in\cP} \{ L_P(\theta) \}$.
This maximizes the loss $L_P$ over a well-defined set of measures $\cP$ that often takes the form
\begin{equation}\label{abscp}
  \cP =\Big\{ P \, | \, D(P, P_{b}) \le \rho , \int dP(\xi) = 1,  P(\xi) \ge 0 \Big\} .
\end{equation}
Here $D(\cdot,\cdot)$ is a metric on the space of probability distributions on $\setX$,
$P_{b}$ denotes a base distribution,
and the constraints limit the feasible candidates to be within a distance $\rho$ of $P_{b}$. 
Our overall goal is to determine the parameters $\theta \in \Theta$ that solve the following DRL problem for a given $\setX$ and $\cP$:
\begin{equation}\label{absfmln}
 R(\theta^*_{rob}) \, = \, \min_{\theta\in\Theta} \, \Big\{ R(\theta) \Big\} \, = \,
    \min_{\theta\in\Theta} \, \Big\{ \sup_{P\in\cP} \{ L_P(\theta) \} \Big\}.
\end{equation}

To solve the DRL problem~\eqref{absfmln}, 
a finite training dataset is used to calculate the optimal values of $\theta^*$ for the model parameters $\theta$,
and then these model parameters are
applied for inference over a separate testing dataset.
Assuming the training and testing datasets to be identically distributed, one can obtain the values $\theta^*_{erm}$ that minimize the empirical loss $L_{U_N}(\cdot)$ under
the equal-weight empirical distribution $U_N := \{1/N\}$ over the training dataset of size $N$.
Unfortunately, use of the values $\theta^*_{erm}$ can result in poor generalization in practice~\citep{perlich03}.
To address this issue, a different approach has been proposed,
which we now briefly summarize (see, e.g., \cite{bwm16,gk17,lm17,nd17,gsw19}).
Note that this approach to solve~\eqref{absfmln} explicitly deals with the ambiguity in the identity of the true (unknown) distribution $P_0$.
Let $L_{P_0}(\cdot)$ denote the loss under the true distribution $P_0$ (were it known), $\theta^*_0$ denote the minimizer of $L_{P_0}(\cdot)$,
and $\theta^*_{rob}$ denote the minimizer of the loss under the corresponding worst-case distribution $P^*(\theta^*_{rob})$ where $P^*(\theta^*_{rob})\neq P_0$. 
Then, for instances of $L_P$ and a proper setting of $\rho$ in~\eqref{abscp}, \citeN{nd17} show that the performance gap at the solution $\theta^*_{rob}$
is $O(1/N)$ whereas the performance gap at the solution $\theta^*_{erm}$ is the considerably larger $O(1/\sqrt{N})$.
Such a great potential is one of the reasons for increased interest in DRL-based approaches for generalization.

Our study concerns efficient solutions of~\eqref{absfmln} as a fundamental approach to generalization,
starting with the inner maximization problem of the 
formulation.
Previous work \citep{bwm16,gao2018,snd18,chen-paschalidis2018,esfahani2018} has considered efficient solutions
of this 
problem;
e.g., explicit solutions are available in special cases of Wasserstein distance constraints that render an explicit characterization of the objective function
$\E_{P^*(\theta)} [l(\theta,\xi)]$~\citep{bwm16,snd18}.
It is important to note, however, that such reductions do not hold in general for all interesting Wasserstein distance metrics~\citep{bk20} and most cases require solving
a convex non-linear program. 
In this paper we therefore consider
the general $\phi$-divergence class of distance metrics:
$D_{\phi}(P,P_b) = \E_{P_b} [\phi( \frac{dP}{dP_b} )]$,
where $\phi(s)$ is a non-negative convex function such that $\phi(s)=0$ only at $s=1$;
e.g.,
the modified $\chi^2$-metric is given by $\phi(s) = (s-1)^2$ and Kuhlback-Leibler divergence is given by $\phi(s) = s\log s - s +1$. 
Define the $N$-dimensional vector variable $P := (p_n)$, and let us set the base distribution $P_b = U_N$.
Hence, the loss function and constraint set $\cP$ are given by $L_{P}(\theta) = \sum_{n=1}^N p_n l(\theta, \xi_n)$ and
$\cP = \{ P \, | \, \sum_{n=1}^N p_n = 1, p_n\ge 0,\forall n, D_{\phi}(P,U_{N}) = \frac{1}{N} \sum_{n=1}^N \phi(Np_n) \le \rho \}.$

Assuming the loss functions $l(\cdot,\xi_n)$ are convex, then this DRL problem~\eqref{absfmln} is convex in $\theta$.
Previous work by
\citeN{bhwmg13} therefore consider
applying classical
SGD
methods to compute the solution of the equivalent dual formulation for this convex-concave case.
After observing that such SGD can become unstable under certain conditions,
\citeN{nd16}
propose an alternative approach (for convex losses $l(\theta,\xi)$) that interleaves an SGD step in each of the $\theta$ and $P$ variables,
where the primal-dual steps apply stochastic mirror-descent to each variable.
However, as shown in~\cite{gsw19}, this algorithm is computationally prohibitive even for relatively small datasets,
with the end results failing to reach the optimal solution.
Similar primal-dual approaches have gained attention in the optimization literature (see, e.g., descent-ascent~\citep{ljj20} and extra-gradient~\citep{mr20} among others)
as general purpose minimum-maximum solvers that solve~\eqref{absfmln} in the combined $d+N$ space of $(\theta, P)$.
Our approach specializes to the DRL formulation and is able to provide a convergent algorithm that performs (stochastic) gradient descent only in the $\theta$ space.

%
In this vein, \citeN{nd17}
propose to directly find the optimal $P^*(\theta)$ that defines $R(\theta)$;
that is, to solve \refeq{absfmln} as a large deterministic gradient descent problem applied to the primal outer variables $\theta$.
They show that 
the inner maximization can be reduced to two one-dimensional root-finding problems for the modified $\chi^2$ case.
This can be solved using bisection search which requires an expensive $O(N\log N)$ amount of time
at each iteration; see Section~\ref{sec:algo}.

\citeN{gsw19} seek to alleviate the computational burden of this costly full-gradient method by applying SGD to the outer variables $\theta$
using stochastic small-sample (or mini-batch)
gradient estimates of the robust loss $R(\theta_t)$ at each iteration $t$.
While subsampling (mini-batching) works well in empirical loss minimization because sample estimates of gradients are unbiased,
\citet[Theorem 3]{gsw19} show that subsampling for DRL induces a bias in the estimation of the (inner) robust loss gradient.
This bias can only be reduced if the subsample size grows, vanishing only when the subsample size $M(t)$ equals $N$~--~i.e., the expensive full-gradient case.
\citeN{gsw19} further show that this tradeoff between bias and computation time can be optimally balanced by a well-chosen growth of $M(t)$ towards $N$,
and that the growing mini-batch SGD method they propose converges without requiring the convexity of $l(\cdot,\xi_n)$.
At the same time, great opportunities remain to improve the efficiencies of solving such large-scale DRL problems.

In this paper we devise
a new small-sample stochastic estimator of the gradient of the robust loss $R(\theta)$ that is unbiased; see~\refprop{zerobias}.
Our construction of this estimator is based on
a novel approach called Multi-level Monte Carlo that has been recently introduced in the research literature by~\cite{giles08}
to eliminate the bias exhibited in typical numerical methods for stochastic problems.
This technique has been adapted to various unconstrained stochastic optimization contexts by~\citeN{bg15} and~\citeN{bgilz17}.
The key idea, in comparison to the increasing $M(t)$ sequence introduced by~\citeN{gsw19}, is to randomize the choice of $M(t)$ in each iteration $t$ in a manner that allows for $M(t)=N$ with a small probability.
While this potentially incurs a high computational cost and the randomized $M(t)$ imply a possibly larger variance for the gradient estimator,
our careful analysis
in Theorems~\ref{smalleffort} and~\ref{smallvar}
addresses these two competing concerns.
Moreover, such results provide the range of randomization parameter values that balances these competing objectives,
leading to an efficient unbiased estimator of $\nabla_{\theta} R(\theta)$.
We then establish convergence of an SGD algorithm with this gradient estimator by exploiting standard tools; see~\refthm{thm:cvg}.

\cite{bk20} use the Multi-level Monte Carlo
idea to approximate solutions to the DRL problem~\eqref{absfmln} with Wassterstein distance.
Like our approach here, the structure inherent in the resulting inner maximization problem allows a transformation to a more tractable but approximate form.
The gradients of this approximation take the form of ratios of expectations over the full dataset, which are expensive to compute exactly.
\cite{bk20} utilize a version of the~\citeN{giles08} algorithm devised for this specific estimation problem and find solutions to their approximate problem with the unbiased estimates of its gradients.
In strong contrast, our algorithm proposes a novel approach that exploits
Multi-level Monte Carlo
to estimate the gradient of $R(\theta)$ based on the bias result in~\refthm{thm:bias},
and our method converges to the solution of the original $D_\phi$-DRL problem.

We conduct a broad collection of numerical experiments that
consider strongly convex DRL formulations of binary classification problems,
comparing the performance of our algorithm against those proposed by~\citeN{nd17} and~\citeN{gsw19}.
The experiments show that our algorithm maintains a performance level similar to these methods.
It is able to do so with orders of magnitude less computation than~\citeN{nd17}, and significant computational improvements over~\citeN{gsw19} for larger datasets which grows with the dataset size.
We further demonstrate that the robust formulation~\eqref{absfmln} can improve on generalization performance over a model trained with the ERM
formulation using regularization tuned via $k$-fold cross validation.
In particular, our experiments show that the same or better generalization guarantee can be obtained with remarkable savings in computational effort.

Lastly, we note that our general approach is also being applied to instances of~\eqref{absfmln} that are formulated with general non-linear convex Wasserstein distance constraints as part of related work,
though not discussed further herein.
We refer to the appendix for all proofs and additional theoretical and empirical materials.

We note that an earlier version of this paper previously appeared in~\cite{gs20}, with the primary difference being the inclusion of comparisons between our Multi-level Monte Carlo approach and $k$-fold cross validation
as presented in Table~\ref{table:gen} of Section~\ref{sec:expt}.


\section{DRL Algorithm} \label{sec:algo}

We present our subgradient descent algorithm for efficiently solving the general DRL minimum-maximum optimization problem in \refalg{alg:overall}.
Consistent with previous work in the literature~\citep{nd16,gsw19},
it follows SGD-like iterations for the outer minimization problem in~\eqref{absfmln}:
\begin{equation}\label{ssgd}
\theta_{t+1} = \theta_t - \gamma_t G_{t}(\theta_t), 
\end{equation}
where $\gamma_t$ is the step size
(or learning rate)
and $G_{t}(\cdot)$ is a stochastic approximation of the gradient of the robust loss $R(\cdot)$ from the
inner maximization formulation over $D_{\phi}$-constrained $\cP$.
This differs from the convex-concave formulations of~\citeN{bhwmg13} and allows us to consider non-convex losses $l$,
as long as $G_t(\cdot)$ approximates the gradient $\nabla_{\theta} {R}(\cdot)$ sufficiently well.
The existence of $\nabla_{\theta} R(\theta)$ can be shown to follow from Danskin's Theorem;
see~\refprop{prop:rgrad} below.
Similar to~\citeN{gsw19},
we also depart
from~\citeN{nd17}
in that we construct the estimate $G_t$ by solving the inner maximization problem restricted to
subsets $\cM$
of the full dataset,
where the subsets ${\cM}$ are created by uniformly sampling without replacement
$M = |\cM|$ values from the complete training dataset of size $N$.
A fundamental difference between our approach and that of~\citet{gsw19} concerns the manner in which this subsampling is performed,
together with their associated tradeoffs.
Since $\phi$-divergence metrics assess distances between probability measures defined on the same support,
we exploit subsampling of the discrete training dataset and the corresponding tools of probability.
\begin{algorithm}[!htp]
	\caption{Zero-Bias Sampled Subgradient Descent}\label{alg:overall}
	{\bf Input}:
Step sizes $\gamma_t$;
	Sampling parameter $r$;
	Initial iterate $\theta_0$.
	\begin{algorithmic}[1]
		\FOR { $t=1,2,\ldots$}
		\STATE\label{step:enstart} 
		Sample $\tau_t$ from truncated geometric $q_k$  
		\STATE Sample $M(\tau_t) + 1$ indices
		{\em without replacement} uniformly from
		$[N]$,
and constitute $\cM(\tau_t)$
		\STATE Solve problems~\eqref{restrob} to obtain gradient~\eqref{subprob:grad} estimates $\nabla_{\theta}\hat{R}_{\tau}(\theta)$, $ \nabla_{\theta}\hat{R}_{\tau,l}(\theta)$, $\nabla_{\theta}\hat{R}_{\tau,r}(\theta)$, $\nabla_{\theta}\hat{R}_1(\theta)$
		\STATE Set $\Delta_{\tau_t}(\theta_t)$, $G_t(\theta_t)$ from~\eqref{deltadefn}, \eqref{gilesdefn}, respectively
		\STATE Set $\theta_{t+1} \gets \theta_t - \gamma_t G_t(\theta_t) \quad$
		\STATE Increment $t\gets t+1$
		\ENDFOR
	\end{algorithmic}
\end{algorithm}

Now, we present the details of our DRL~\refalg{alg:overall}, starting with general aspects of the subsampling approach consistent with~\citeN{gsw19}.
Define $[N] := \{1,\ldots,N\}$, ${P} := ({p}_m)$ of dimension $M$, and objective coefficients $z_m := l(\theta,\xi_m)$.
We consider
\begin{align} \label{restrob}
&\hat{R}_{M}(\theta) \quad=\quad \max_{{P}=({p}_m)} \sum_{m\in\cM}{p}_m z_m  \\
&\mbox{s.t. }  \sum_{m\in\cM} \phi(M {p}_m) \le M\rho_M,\, \, \sum_{m\in\cM} {p}_m = 1 ,\, {p}_m \ge 0,  \nonumber
\end{align}
where the constraint parameter $\rho_M$ adds a small inflation term to $\rho$ that changes with $M$; see~\refthm{thm:bias} below.
Define ${P} = ({p}_m)$ of dimension $M$ and write the Lagrangian objective of~\refeq{restrob} as
\begin{align}
&\cL(\theta, \alpha,\lambda, {P})
= \sum_{m\in\cM} l(\theta,\xi_m) {p}_m + \lambda \bigg(1 - \sum_{m\in\cM}{p}_m \bigg) \;\; \nonumber \\
&\qquad\qquad\qquad+ \frac{\alpha}{M} \bigg(M  \rho_{M} - \sum_{m\in\cM}\Phi(M{p}_m) \bigg) .
\label{roblag}
\end{align}
We then have the optimal objective value 
$\hat{R}^*_{M}(\theta) = \min_{\alpha\ge 0,\lambda} \max_{\hat{p}_m\ge 0} \cL(\alpha, \lambda, {P})$~\citep{lbgr69}.
In addition,~\refprop{prop:rgrad}
shows how a valid gradient for $\hat{R}_M(\theta)$ is obtained and also
includes imposed conditions that prevent any degenerateness in~\eqref{restrob}.
Henceforth
we shall assume that
${P}^*(\theta) = ({p}^*_{m}(\theta))$ is an optimal primal solution to~\eqref{roblag}. 
%
\begin{proposition}[{\citet[Proposition 1]{gsw19}}]\label{prop:rgrad}
Define $\Theta_{\varnothing} \DefAs  \{\theta : l(\theta, \xi_{n_1}) = l(\theta, \xi_{n_2}),\,\,\forall\,n_1,n_2 \}$.
For a small $\varsigma>0$, further define $\Theta_{\varnothing,\varsigma} \DefAs \cup_{\theta_o \in \Theta_{\varnothing} } \{\theta : |\theta - \theta_{o}| < \varsigma \}$
to be the $\varsigma$-neighborhood of $\Theta_{\varnothing}$.
Let the feasible region $\Theta$ be compact and assume $\Theta\subseteq \Theta_{\varnothing,\varsigma}^c$.
Further suppose $\phi$ in the $D_{\phi}$-constraint has strictly convex level sets,
and let $\rho <
\bar{\rho}(N,\phi) = (1-\frac 1 N) \,\phi(\frac  N {N-1}) + \frac 1 N \phi(0)$.
	Then, (i) the optimal solution $P^*$ of $R(\theta)= \sup_{P\in\cP} \{L_P(\theta)\}$ is unique, and (ii) the gradient is given by
$\nabla_{\theta} {R}(\theta) = \sum_{n\in\cN}{p}^*_{n}(\theta) \,\nabla_{\theta} l(\theta,\xi_n)$.
\end{proposition}


Then, given the solution ${P}^*(\theta)$ to~\eqref{roblag}, a valid subgradient for $\hat{R}_t(\theta_t)$ is obtained from \refprop{prop:rgrad}(ii) as
\begin{equation}
\nabla_{\theta} \hat{R}_{M}(\theta) \quad=\quad
\sum_{m\in\cM}{p}^*_{m}(\theta) \,\,\,\,\nabla_{\theta} l(\theta,\xi_m).
\label{subprob:grad}
\end{equation}
\refthm{thm:bias} below provides a bound on the error in using \eqref{subprob:grad} to approximate the true full-sample robust loss gradient $\nabla_{\theta}{R}(\theta)$ as a function of the sample size $M$.
This bound on the bias vanishes only as $M\uparrow N$ as the iterate $t\tndi$.
Since fixed bias violates a basic requirement for SGD that the gradient estimator
$\E[\nabla\hat{R}(\theta)] = \Theta(\nabla R(\theta))$\footnote{Standard complexity notation; see, e.g.,~\citeN{sipser06}.}
(\citet[Section~4.3]{bcn16}),
then the convergence of \eqref{ssgd} cannot be guaranteed when
$M$ is fixed for all iterations $t$ and $M<N$.

In order to eliminate bias,
the algorithm of~\citet{gsw19} needs to increase its sample size $M$ toward $N$ as the iterations $t$ grow, consequently also increasing the computation time.
The
maximum size $N$ is then hit after a large
number of iterations $T$, at which point their algorithm switches to the deterministic optimization of~\citeN{nd17}.
\citet{gsw19} establish the convergence of their algorithm and show how growth parameters of the per-iterate samples $M(t)$ can be carefully chosen to minimize the overall computation time
needed to converge to the optimal solution $\theta^*_{\rob}$. 

We propose here
a fundamentally different approach
that eliminates bias without the concomitant increase in computation by adding a Multi-Level Monte Carlo randomization step;
see~\citet{giles08} and \citet{bg15} for a basic introduction to Multi-Level Monte Carlo randomization.
For ease of exposition, we henceforth assume the training set
is
such that $N=2^K$ for an integral value of $K$,
noting that the general case is easily handled by appropriately adjusting the algorithm below.
Let $\tau$ be a discrete random variable taking values in $[K]$, where $K=\log N / \log 2$. The random variable $\tau$ is sampled geometrically using the probability mass function
$$q_k \DefAs P(\tau = k) = r^{k-1} \frac {1-r}  {1-r^{K+1}},\quad k=1,\ldots,K.$$ 
Let $M(\tau)=2^{\tau}$ be a subset size associated with $\tau$, and $\cM(\tau)$ the corresponding subset sampled uniformly without replacement from the training dataset.
Partition $\cM(\tau)$ into two equal-sized subsets $\cM_l(\tau)$ and $\cM_r(\tau)$, each of size $2^{\tau-1}$.
To simplify notation,
we
denote
the robust loss calculated over $\cM(\tau)$, $\cM_l(\tau)$ and $\cM_r(\tau)$ by $\hat{R}_{\tau}$, $\hat{R}_{\tau,l}$ and $\hat{R}_{\tau,r}$, respectively.
Define
\begin{align}
\Delta_\tau(\theta) &\DefAs \nabla_{\theta}\hat{R}_{\tau}(\theta) - \frac { \left( \nabla_{\theta}\hat{R}_{\tau,l}(\theta)+ \nabla_{\theta}\hat{R}_{\tau,r}(\theta)\right)} 2 \label{deltadefn}\\
G(\theta) &\DefAs \nabla_{\theta}\hat{R}_1(\theta) + \frac {\Delta_{\tau} (\theta)} {q_{\tau}},\label{gilesdefn}
\end{align}
where $\nabla_{\theta}\hat{R}_1(\theta)$ is the gradient computed from a singleton subset, as per the definition~\eqref{restrob} with $M=1$.
Then the estimator $G(\theta)$ provides the unbiased estimator for $\nabla_{\theta} R(\theta)$, in the style of~\citet{giles08},
under very general conditions on $l(\cdot, \xi)$. 
In fact, this allows our algorithm below to be used in important cases when $l(\cdot,\xi_n)$ are non-convex, such as training deep learning models. 


Next, we consider an exact solution to the inner maximization problem.
In particular, we obtain the optimal primal and dual variables for various $\phi$ functions by
exploiting a general procedure to solve Lagrangian formulations.
This procedure, presented as \textbf{IM Procedure} in the appendix, is also used by~\citeN{gsw19}
who take a basic approach for such problems pursued by others (see, e.g.,~\citeN{bhwmg13,gl18,nd16,nd17}) and adapt it to the inner maximization
formulation over $D_{\phi}$-constrained $\cP$.
\citet[Proposition 2]{gsw19} establish a worst-case computational complexity bound of $O (M \log M + (\log (\frac{1}{\epsilon}) )^2)$
for finding an $\epsilon$-optimal solution to~\eqref{restrob} when applying \textbf{IM Procedure} to any $\phi$-divergence.
Since the machine-precision $\epsilon$ is set to a fixed arbitrarily small value independent of any other parameter of the formulation or algorithm (e.g., $M, N,\rho$),
we follow \citeN{gsw19} and assume that \textbf{IM Procedure} returns the exact unique solution $(P^*,\alpha^*,\lambda^*)$ to~\eqref{restrob} with computational
complexity bounded by $O(M\log M)$.

Now, we consider the bias induced by the approach of subsampling the full support using the subgradient approximation $\nabla_{\theta}\hat{R}_M(\theta)$ to the true gradient $\nabla_{\theta}R(\theta)$.
%
Let $P^*_M = (p^*_1,\ldots,p^*_{M})$ denote the optimal solution of~\eqref{restrob} when restricted to the set $\cM$ of size $M$,
and let $P^*=(p^*_1,\ldots,p^*_N)$ denote the optimal solution to the full-data version of~\eqref{restrob}.
Further let $\E_{M}$ and $\Prb_{M}$ denote expectation and probability with respect to the random set $\cM$, respectively.
The following
result
shows
the induced bias to be of order $O(1/M - 1/N)^{1-\delta}$ for an appropriate $\delta>0$.
\begin{theorem}[{\citet[Corollary 1]{gsw19}}]\label{thm:bias}
  Suppose the $\phi$-divergence satisfies uniformly for all $s$ and $\zeta<\zeta_0$ the continuity condition $|\phi(s(1+\zeta))-\phi(s)| \le \kappa_1 \zeta \phi(s) + \kappa_2 \zeta$, for constants $\zeta_0, \kappa_1, \kappa_2>0$.
  Further suppose the assumptions of~\refprop{prop:rgrad} hold.
  Define $\eta_{M} = c (\frac{1}{M} -\frac{1}{N})^{(1-\delta)/2}$ for small constants $c, \delta >0$, and set the $D_{\phi}$-target in~\eqref{restrob} to be $\rho_{M} = \rho + \eta_{M}$.
	Then, $\| \E_{M} [\nabla\hat{R}_{M}(\theta)]  - \nabla{R}(\theta)\|^2_2 = O(\eta_{M}^2).$
\end{theorem}
Theorem~\ref{thm:bias} shows that a squared bias of order $\eta^2_M\approxeq M^{-1}$ is incurred when a fixed subsample of size $M$ is used in creating the gradient estimate.
This motivated the approach of~\citeN{gsw19} to progressively reduce the bias.
However, our alternative approach based on the Giles estimator~\eqref{gilesdefn} completely eliminates this bias using the additional random variable~$\tau$.

\section{Theoretical Analysis} \label{sec:analysis}
We next establish various mathematical properties for our DRL approach, including properties of $G(\theta)$ and convergence.
This analysis provides theoretical justification for our algorithm and its parameter settings.

To start, we can easily establish the unbiasedness of the estimator $G(\theta)$ given in \eqref{gilesdefn}.
Let $\E_{\tau}$ denote expectation under the joint distribution of the randomizing variable $\tau$ and the subsampled set $\cM(\tau)$.
\begin{proposition}\label{zerobias}
	We have that $E_{\tau} [G(\theta)] = \nabla_{\theta}R(\theta)$.
\end{proposition} 

Observe that the computation time in sampling the zero-bias estimator now depends on the randomizer $\tau$ because we solve inner maximization problems of size $M(\tau)=2^{\tau}$.
Let $T$ denote the total computation time involved in constructing one replication of the Giles estimator $G(\theta)$. 
We then establish the following result.
\begin{theorem} \label{smalleffort}
	When $r< 1/2$, the expected computation time is bounded by a quantity independent of the dataset size $N>1$ as follows:
	$$\E_{\tau}[T]  \le \frac {4C' r(1-r) \log 2} {(1-2r)^2} ,$$
	where constant $C'$ is finite.
\end{theorem}

It is also important to ensure that the injection of extraneous randomness via $\tau$ does not result in a very large variance of the estimator $G(\theta)$.
We establish that the variance of our Giles estimator is bounded 
when the variance of a fixed $M$ gradient estimator is of the same order as exhibited by the sample average \citep{wilks} of an $M$-sample set
chosen uniformly without replacement from a large finite set of size $N$,
and therefore $O(\frac 1 {M} - \frac 1 N) = O(\eta_{M}^{2/(1-\delta)})$. 
%

\begin{theorem} \label{smallvar}
Suppose 
the variance of an estimator $\nabla \hat{R}_M(\theta)$ with subsample size $M$ obeys  $\E \left[ \left\|\nabla\hat{R}(\theta) -
\E [\nabla\hat{R}(\theta)]\right\|^2_2 \right] \le   
C\left(\frac 1 {M} - \frac 1 N\right)$.
Then, when $r\in (1/4, 1/2)$, the variance of $G(\theta)$ is bounded from above by a quantity independent of the dataset size $N>1$ as follows:
	$$\E_{\tau} \left[\|G(\theta)-\nabla_{\theta} R(\theta) \|^2_2 \right]  \le C(r)<\infty. $$  
\end{theorem}
A complete expression for the constant $C(r)$ is included in the proofs provided in the appendix.



Now, we turn to analyze the convergence of \refalg{alg:overall}, 
for which a common requirement is that the objective $R(\theta)$ be Lipschitz smooth.
The gradient $\nabla_{\theta}R(\theta)$ of the optimal value of an optimization problem is in general not Lipschitz if the objective function is Lipschitz.
Consider, for example, a linear objective $l(\theta,\xi_n) = \theta^t\xi_i$, which implies $R(\theta) = \max_p \sum_i p_i \theta^t\xi_i$.
When maximized over a {\em polyhedral} constraint set (e.g., the probability simplex constraints of~\eqref{absfmln})  the $0$-Lipschitzness of the $l$ are not preserved for $R$, because in this case the optimal solutions $P^*$ are picked from the discrete set of vertices of the polyhedron and thus $\nabla_{\theta}R(\theta)$ is piecewise discontinuous.

Our assumptions from~\refprop{prop:rgrad} yield an inner maximization with a non-zero linear objective over a \textit{strictly convex} feasible set.
The desired smoothness can then be obtained with some additional conditions on the loss functions $l(\theta,\xi_i)$.
\cite{gsw19} note that 
the Lipschitzness of $\nabla_{\theta} R(\theta)$ follows from the Hessian of $R(\theta)$ being bounded in norm, which is often satisfied by common statistical learning losses such as log-logistic and squared losses of linear models over compact spaces.

The combination of the Lipschitzness of the robust loss function $R(\theta)$,
the finiteness of the variance of the Giles gradient estimator $G(\theta)$,
and the finite expected computation time to obtain the estimator enables us to now apply the standard SGD convergence machinery in establishing the convergence of~\eqref{ssgd} to first-order optimal solutions.
We exploit the following 
result
adapted to our problem and its setting. 
\begin{theorem}[{\citet[Theorem 4.9]{bcn16}}]\label{thm:cvg}
	Suppose the robust loss objective $R(\theta)$ satisfies:
%
(i) A lower bound $R_{\inf}$ exists for the robust loss function $R(\theta)\ge R_{\inf}$, $\,\,\forall \theta\in\Theta$;
		\noindent (ii) The gradient $\nabla_{\theta} R(\theta)$ is $L$-Lipschitz.
	Further suppose the estimator $G(\theta)$ of the gradient $\nabla_{\theta}R(\theta)$ is unbiased and has variance bounded above by a constant $C<\infty$.
	Choosing the step size sequence $\gamma_t$ to satisfy $\sum_t \gamma_t \tndi$ and $\sum_t \gamma_t^2 < \infty$, we then have
	$$ \liminf_{t\tndi } \E \left[ \|\nabla_{\theta} R(\theta_t)\|^2_2 \right] =0.$$
\end{theorem}

\section{Numerical Experiments}
\label{sec:expt}
In this section we present a broad spectrum of numerical experiments conducted to empirically evaluate our new DRL approach and its theoretical properties presented above.
Section~\ref{ssec:alg} considers empirical results that compare our approach against previous work in the research literature,
whereas Section~\ref{ssec:gen} delves into the generalization power observed for our approach.

\subsection{Algorithmic Performance}\label{ssec:alg}
A large collection of numerical experiments were conducted to empirically evaluate our new unbiased Giles estimator sampled subgradient descent algorithm, denoted as GSSG,
in comparison with the full-support gradient algorithm of~\citeN{nd17}, denoted as FSG, and the progressively sampled subgradient descent algorithm of~\citeN{gsw19}, denoted as PSSG.
Following the previous work in this research literature, all experiments fall within the context of training binary classification models.
We also include the standard SGD (i.e., fixed minibatch $M(t)=M$) method to estimate the impact of bias identified in~\refthm{thm:bias}.
All of these methods were re-implemented since full source code was not made available with the corresponding
publications.\footnote{We will open our github repository for public consumption once the anonymity requirement is resolved.}

Each numerical experiment uses a logistic regression loss function $l(\theta; (x,y)) = \log (1 + \exp(-y \theta^t x))$,
where $x$ represents an $N \times d$ matrix with $N$ the total number of samples and $d$ the number of features,
and where $y$ represents the binary $N$-dimensional class labels encoded as $\pm 1$.
In every experiment, a large dataset was split to randomly select $25\%$ of the data for the testing dataset with the remaining $75\%$ of the data used for the training dataset.
The two
main representative datasets considered in this section are the HIV-1 Protease Cleavage
dataset ($N\sim 10^4$ samples) from~\cite{UCI},
and the Reuters Corpus Volume I dataset ($N\sim 10^6$ samples) from~\cite{RCV1}.

\begin{figure}[htbp]
	\vskip -0.15in
	\begin{center}
		\includegraphics[width=0.6\columnwidth]{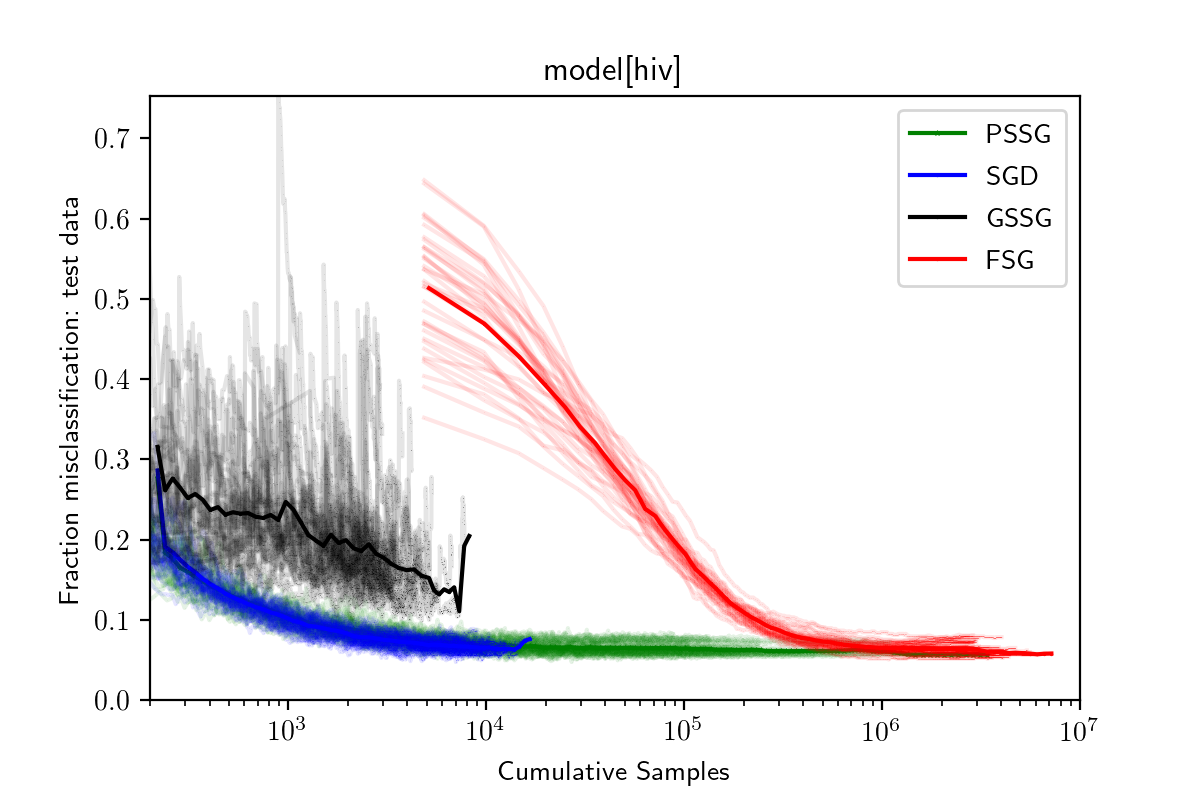}
		\caption{{\small\textit{Comparison of PSSG (green), FSG (red), GSSG (black) and standard SGD (blue) on misclassification in testing as a function of cumulative samples over HIV-1 dataset with $\rho=0.1$ and log-scale $x$-axis.}}}
		\label{fig:rho0.1:HIV}
	\end{center}
	\vskip -0.175in
\end{figure}

\citeN{bwm16} and \citeN{nd17} consider at length the question of setting the parameter $\rho$ in~\eqref{abscp}, providing a broad guideline for binary classification with 
logistic loss models that
$\rho = O(\sqrt{d/N})$.
For our datasets, this suggests that $\rho$ range from $O(0.01)$ to $O(1)$, and therefore we consider $\rho=0.1$ in the main body of the paper and provide additional results for $\rho$ across a wider range 
in the appendix.
In all cases, the inner-maximization formulation is solved to within $\epsilon$-accuracy ($\epsilon = 10^{-7}$ in our experiments).
For consistency with the primary results presented in~\citep{gsw19}, we present our results as a function of the cumulative sample size;
the corresponding results over wall-clock times are presented in the appendix.

Parameter $\delta$ appears prominently in the definition of $\rho_M$ and in defining the bias of the gradient estimation in~\refthm{thm:bias}, but the result requires only that
$\delta$ be a small positive constant.  We find that the numerical experiments
are not sensitive to $\delta$ and therefore present below results
for $\delta=0.01$ in defining the expanded constraint $\rho_M$.
The standard SGD method and our GSSG method both require a diminishing step-size sequence, which we choose to be $\gamma_t= 5000/(5000+t)$.
The \emph{constant-growth factor} for PSSG is set to $\nu = 1.001$, as in~\cite{gsw19}.
The GSSG estimator parameter $r$ has a feasibility range of $(1/4,1/2)$.
\citeN{bg15} consider a Giles-based estimator for unconstrained optimization problems with continuously distributed random variables, while we solve the constrained problem~\eqref{restrob} using discrete samples from the finite training set.
They show that the product $\E \{T\} \times \var \{ G(\theta)\}$,
denoting the variance of a generalized central limit obeyed by the Giles estimator as a function of the computational budget,
is minimized by $r^*=2^{-3/2}$ which represents the geometric mean of the end points of the feasible region.
We therefore use this choice in the experiments for our GSSG method.


\begin{figure}[htbp]
	\vskip -0.15in
	\begin{center}
		\includegraphics[width=0.6\columnwidth]{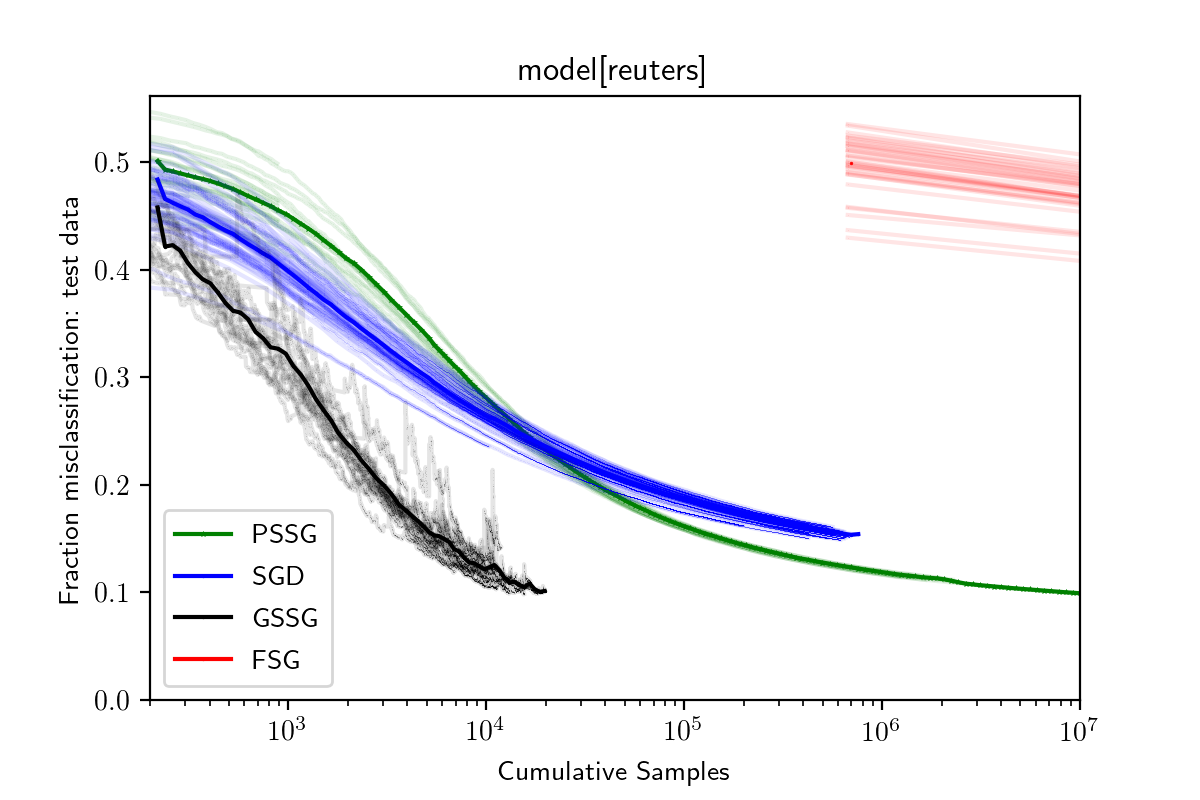}
		\caption{{\small\textit{Comparison of PSSG (green), FSG (red), GSSG (black) and standard SGD (blue) on misclassification in testing as a function of cumulative samples over RCV1 dataset with $\rho=0.1$ and log-scale $x$-axis.}}}
		\label{fig:rho0.1:RCV1}
	\end{center}
	\vskip -0.2in
\end{figure}

\subsubsection{HIV-1 Protease Cleavage}    
\label{sssec:HIV}
The HIV-1 protease cleavage dataset helps develop effective protease
cleavage inhibitors by predicting whether the HIV-1 protease will
cleave a protein sequence
in its central position (coded as class-label $1$) or not (class-label $-1$).
We preprocessed the data to
remove conflicting and overlapping samples, following
\cite{Rogn}.
This leaves a total of $5830$ samples of $(d=160)$-dimensional feature vectors using orthogonal binary representation.
These samples have $991$ cleaved and $4839$ non-cleaved labels.
We sampled the initial $\theta_0$ from $U[-1,1]$, and again set $\rho=0.1$.
%

Figure~\ref{fig:rho0.1:HIV} compares the average test missclassification loss of the four algorithms.
We observe that PSSG is significantly faster than FSG and the effect of bias on SGD is muted.
The SGD algorithm essentially follows down the same path as the PSSG algorithm.
On the other hand, our GSSG algorithm seems to struggle somewhat in comparison.
This illustrates one side of the tradeoff between the bias that the standard SGD may be subject to versus the higher variance experienced by our GSSG algorithm. Here the HIV-1 case is small enough that the bias is not noticeable but the added variance of the Giles estimator due to the added randomization adversely affects our GSSG algorithm.
On the computational side, the size of the dataset is small enough that the PSSG method with increasing subsample size is quick to execute.
Meanwhile, one of our primary interests and motivations in devising the GSSG algorithm concerns datasets of much larger scale and greater complexity, as we consider next.

\subsubsection{Reuters Corpus Volume I}
\label{sssec:RCV}
The Reuters Corpus Volume I (RCV1) dataset comprises $804414$ samples,
with $(d=47236)$-dimensional feature vectors, thus being a
substantially larger and more challenging dataset.
The primary purpose is to classify each sample article as either belonging to a corporate/industrial
category or not based on its content~\citep{RCV1}, coded as $1$ if true and $-1$ otherwise.
We initialized the values of $\theta_0$ to $U[-1,1]$ in every numerical experiment and again set $\rho=0.1$.

\reffig{fig:rho0.1:RCV1} presents our empirical results from $10$ experiments for GSSG, PSSG, FSG and SGD,
comparing their fractional misclassification performance over the testing dataset
as a function of the cumulative sample size used by each algorithm (log scale).
As expected, FSG is simply not competitive with the other methods for this larger problem instance.
The impact of the bias on the SGD method induced by the fixed $M$ is noticeable in this case, resulting in the convergence of SGD to a poorer solution.
Both the PSSG and the GSSG methods find the best generalizing solutions, but in contrast to the results for the smaller dataset above, the added variance
experienced by our GSSG method does not hinder its progress in the case of this much larger and more complex dataset, as expected.
However, the larger subset size in the later iterations of PSSG does slow it down in the case of this much larger dataset, as compared to GSSG. 
Indeed, GSSG converges to the solution significantly faster than PSSG, which in turn converges much faster than FSG, as also illustrated in the CPU time columns of Table~\ref{table:gen}. 

\subsection{Generalization Performance}\label{ssec:gen}
Given the importance of generalization as a goal in machine learning to solve model-fitting with either regularization penalty terms or DRL,
we turn to study the impact of the robust formulation~\eqref{absfmln} on improving generalization of learning models. 
To estimate the power of our DRL approach in this regard, we compare the results of the GSSG method
in solving the DRL formulation against the PSSG method and an ERM-trained model that is regularized via $10$-fold cross
validation ($10$-fold CV). The \emph{$k$-fold CV} procedure partitions
the full dataset into $k$ equal parts and trains a {\em regularized}
model over each dataset formed by holding out one of the $k$ parts as
the test dataset. We repeat this $k$ times to render $k^2$ test comparisons.
The term $\lambda\|\theta\|_2^2$ regularizes the ERM loss objective, and
a fixed-batch SGD (size $10$) is used to fit the model for each combination
of partition and $\lambda$ values.
Enumeration, starting with $10^6$ and backtracking till the \emph{average} performance over the $10$ validation datasets does not improve for three $\lambda$ successive values,
is used to tune for the optimal penalty parameter $\lambda$ over the partitions.

Table~\ref{table:gen} presents results from running GSSG, PSSG and the ERM method, where experiments were conducted over $14$ public-domain datasets from
UCI~\cite{UCI}, OpenML~\cite{openml} and SKLearn~\cite{RCV1}; the dataset sizes range from $O(10^2)$ to $O(10^6)$. 
The table presents a comparison of the test misclassification produced by our GSSG algorithm, the PSSG algorithm of~\citeN{gsw19}, and the regularized ERM algorithm at termination for the $14$ datasets.
The 95\% confidence intervals are calculated over $10$ permutations of the datasets into training (80\%) and testing (20\%) sets.
For each dataset, the method that produces the best generalization error
(within the specified confidence intervals)
is highlighted in bold.
As evident from the table, the DRL formulation (solved with either PSSG or GSSG) produces models of equal or better quality as produced by the regularized ERM formulation in most datasets, and tends to work better with the larger datasets.
%
\begin{table}[thbp]
	\centering
	\vskip -0.1in
	\setlength{\tabcolsep}{2pt}
	\begin{tabular}{l||r:r|r:r|r:r||r|r|r}
	\hline
		\textbf{Dataset}&\multicolumn{6}{|c||}{\textbf{Test Misclassified (\%)}}	
		&\multicolumn{3}{|c}{\textbf{CPUTime (sec)}}\\
		\hline
		&\multicolumn{2}{c|}{GSSG} &\multicolumn{2}{c|}{PSSG} &\multicolumn{2}{c||}{ERM} &GSSG &PSSG	& ERM\\
		\hline\hline
		tr31.wc$^{\dag}$		&$\mathbf{2.9 }$ & ${ 0.5}$	&$\mathbf{2.6 }$ & ${ 0.3}$	&$\mathbf{2.7 }$ & ${ 0.1}$	&9	&\textbf{6} &987\\ 
		hiv1$^{\ast}$			&$\mathbf{6.1 }$ & ${ 0.5}$	&${5.9 }$ & ${ 0.1}$	&$\mathbf{5.6 }$ & ${ 0.0}$	&11	&\textbf{10}	&1012 \\ 
		gina\_prior$^{\dag}$	&${14.4 }$ & ${ 0.8}$	&$ {13.0 }$ & ${ 0.5}$	&$\mathbf{11.8 }$ & ${ 0.1}$ &\textbf{11}	&12	& 1147\\
		la1s.wc$^{\dag}$		&${10.3 }$ & ${ 1.4}$	&$\mathbf{8.3 }$ & ${ 0.2}$	&$\mathbf{8.5 }$ & ${ 0.0}$	&13	&\textbf{12}	&2456 \\
		gina\_agnostic$^{\dag}$ &${14.8 }$ & ${ 1.8}$	&${13.9 }$ & ${ 0.3}$	&$\mathbf{12.6 }$ & ${ 0.1}$ &\textbf{12}	&13	& 1765\\
		bioresponse$^{\dag}$	&${25.6 }$ & ${ 1.4}$	&${24.2 }$ & ${ 0.4}$	&$\mathbf{21.6 }$ & ${ 0.2}$ &\textbf{14}	&16	& 2791\\
		ova\_breast$^{\dag}$	&${4.1 }$ & ${ 0.6}$	&${3.0 }$ & ${ 0.1}$	&$\mathbf{1.8 }$ & ${ 0.1}$	&\textbf{14}	&17	& 4310\\ 
		fabert$^{\dag}$			&${10.4 }$ & ${ 0.4}$	&$\mathbf{9.8 }$ & ${ 0.0}$	&${10.1 }$ & ${ 0.0}$	&\textbf{16}	&20	& 4128\\ 
		dilbert$^{\dag}$		&$\mathbf{1.6 }$ & ${ 0.6}$	&$\mathbf{1.2 }$ & ${ 0.1}$	&$\mathbf{1.1 }$ & ${ 0.0}$ &\textbf{27}	&33	& 8543\\ 
		adult$^{\ast}$			&${17.1 }$ & ${ 0.3}$	&$\mathbf{16.6 }$ & ${ 0.1}$	&$\mathbf{16.7 }$ & ${ 0.0}$ &\textbf{29}	&36	& 2542\\ 
		imdb.drama$^{\dag}$		&${37.1 }$ & ${ 0.7}$	&$\mathbf{36.2 }$ & ${ 0.1}$	& ${37.1 }$ & ${ 0.0}$ &\textbf{74}	&89	& 19436\\
		guillermo$^{\dag}$		&${32.6 }$ & ${ 1.5}$	&$\mathbf{30.2 }$ & ${ 0.5}$	& $\mathbf{30.7 }$ & ${ 0.1}$ &\textbf{89} &116 & 31547\\ 
		riccardo$^{\dag}$		&${2.4 }$ & ${ 0.3}$	&${1.6 }$ & ${ 0.0}$	&$\mathbf{1.5 }$ & ${ 0.0}$	&\textbf{99}	&120 & 86575\\ 
		rcv1$^{\ddag}$			&$\mathbf{5.5 }$ & ${ 0.1}$	&$\mathbf{5.4 }$ & ${ 0.0}$	&${5.6 }$ & ${ 0.0}$	&\textbf{441}	&543 & 701843\\ 
		\hline
	\end{tabular}
\caption{
{\small\textit{
Comparison of the GSSG, PSSG and regularized ERM formulations over $14$ publicly available machine learning (ML) datasets,
from UCI$^{\ast}$ \citep{UCI}, OpenML$^{\dag}$ \citep{openml} and SKLearn$^{\ddag}$ \citep{RCV1},
arranged in increasing training set size.
The data for PSSG and ERM are reproduced from~\citeN{gsw19}.
The first set of columns provides a 95\% confidence interval of the percentage misclassified over withheld test datasets, with the best-performing method highlighted in bold.
The second set of columns provides the average CPU time taken to solve each formulation.}
}}
\label{table:gen}
\vskip -0.2in
\end{table}

Recall that both GSSG and PSSG solve a single instance of the DRL formulation~\eqref{absfmln} to provide this level of performance, thus avoiding the burdensome $10$-fold CV enumeration.
Table~\ref{table:gen} also provides the average CPU time in seconds recorded over the $10$ permutations.
The average time taken by the ERM $10$-fold regularization, in solving its formulation multiple times in a serial computing mode to identify the best $\lambda$,
exceeds that taken on average to solve the DRL formulations by two to three orders of magnitude.
(We note that the computational benefits of DRL would be even larger if all $\lambda$ values over all grid points were enumerated; see the appendix.)
The GSSG algorithm also provides a reduced computation time in comparison to the PSSG algorithm, with the difference growing with the size of the data set;
the rows in Table~\ref{table:gen} are ordered in increasing size of the training datasets.
As discussed earlier, this is a result of the tradeoff between the higher variance experienced by each GSSG iteration and the larger sample sizes necessitated by the PSSG algorithm,
where the tradeoff tips in the favor of GSSG and grows as the dataset size increases.

The computation time of a single DRL run (GSSG or PSSG) is of the same order as that of a single ERM run.
This indicates a significant computational savings in using DRL because of the elimination of the expensive hyper-parameter tuning step.
The optimal $\lambda$ required by the regularized ERM over the $14$ datasets has a significant amount of variability with no evident pattern relating to dataset characteristics, 
which highlights the need for ERM computations over a wide range of $\lambda$ values for each dataset.
The performance of the DRL formulation requires an adequate choice of $\rho$ (set to $0.1$ here),
and the results for
GSSG (and PSSG)
can
be further improved by a very small number of additional runs with different $\rho$ values while still maintaining significant computational benefits over ERM;
see the appendix.

\subsection{Summary}
Our empirical results support our theoretical results and show that GSSG provides the same quality of performance as PSSG of~\cite{gsw19} and as FSG of~\cite{nd17},
but with orders of magnitude less computation time as compared to FSG and the ERM method.
The GSSG also outperforms the (biased gradient) SGD method for the same step length sequence as a consequence of the bias elimination correction of GSSG.
Our results further show that the DRL formulation renders models of comparable or better quality as those from regularized ERM but with the aforementioned significant computational improvements,
and moreover GSSG improves the computation time over PSSG.
We observe this advantage to be more significant as the dataset size grows.
Hence DRL provides a strong alternative machine learning approach to improve model generalization, especially with GSSG for large dataset size scaling.

\clearpage

\bibliography{robust_sgd}
\bibliographystyle{abbrvnat}

\appendix

\clearpage
\newpage
\newpage
\newpage

\newpage

\appendix
\section*{Appendix: \textit{Zero-Bias Estimation for Stochastic Gradient Descent in Distributionally Robust Learning}}
In this appendix, we present additional theoretical and empirical results together with additional technical details.
Section~\ref{apdx:theory} provides more information on aspects of our DRL algorithm and our theoretical analysis,
including all proofs of our theoretical results, supporting theoretical results, and associated details.
Section~\ref{apdx:expt} provides additional empirical results from our numerical experiments and all associated details.

\section{Algorithm and Analysis}\label{apdx:theory}
%

\subsection{DRL Algorithm}
We first consider solving the linear program~\eqref{restrob} and the Lagrangian formulations~\eqref{roblag}.
To preclude the possibility of degenerate optimal solutions for some objective coefficients $l(\theta,\xi_i)$ in~\eqref{restrob},
we introduce the following assumption.
\begin{assumption}\label{asm:rhomax}
The parameter $\rho$ in~\eqref{restrob} only admits feasible pmfs that assign non-zero mass to all support points, that is, it satisfies
$\rho < \rho_1 = \left(1-\frac 1 N\right) \,\phi\left(\frac  N {N-1}\right) + \frac 1 N \phi(0)$.
\end{assumption}
For strictly convex functions $\phi(\cdot)$,~\refasm{asm:rhomax} ensures that problem~\eqref{restrob} has a unique optimal solution $P^*$
and
$D_{\phi}(P^*, U_N)=\rho$.
Assumption~\ref{asm:rhomax}
and its consequence will be held for the
rest
of the paper, as noted in the statement of Theorem~\ref{thm:bias}.

We now present the following general procedure to solve the Lagrangian formulations~\eqref{roblag} 
for a given iteration $t$.
This basic approach has been pursued
in previous work such as~\citeN{bhwmg13,gl18,nd16,nd17,gsw19}.\\

\textbf{IM Procedure.}
\begin{enumerate} 
	\item 
	{\em Case:} 
	$\alpha^*=0$ along with constraint $D_{\phi}({P}^*_{t},P_b) \le \rho_t$.
	\begin{enumerate}
		\item Let $\cM'_t = \{m\in\cM(t) : z_m = \max_{u\in\cM(t)} z_u \}$ and
		$M'_t = |\cM'_t|$. Set $\alpha^*=0$ in \refeq{roblag}, and then an optimal solution is ${P}^*$ where ${p}^*_m = \frac{1}{M'_t},\;\forall m\in\cM'_t$,
		and ${p}^*_m =0,\; \forall m\notin\cM'_t$.
		\item If $D_{\phi}({P}^*, P_b) \le \rho_M$, then
		{\bf stop} and return ${P}^*$.
	\end{enumerate}
	\item {\em Case:} constraint $D_{\phi}({P}^*_{t},P_b) = \rho_t$ with
	$\alpha^*\ge 0$.
	\begin{enumerate}
		\item Keeping $\lambda, \alpha$ fixed, solve for the optimal
		${P}^*_t$ (as a function of $\lambda,\alpha$) that maximizes
		$\cL(\alpha, \lambda, {P})$, applying the constraint
		${p}_m\ge 0$.
		\item Keeping $\alpha$ fixed, solve for the optimal
		$\lambda^*$ using the first order optimality condition on
		$\cL(\alpha,\lambda,{P}^*_t)$.  Note that this is equivalent to
		satisfying the equation $\sum_{m\in\cM(t)}{p}^*_m = 1$.
This
step is at worst a bisection search in one dimension, 
for which finite bounds on the range of values to search over $\lambda$ are available,
but in some cases (e.g., KL-divergence) a solution $\lambda^*$ is available in closed form.
		\item Apply the first order optimality condition to the
		one-dimensional function $\cL(\alpha, \lambda^*(\alpha), {P}_t^*)$ to
		obtain the optimal $\alpha^* \ge 0$. This is equivalent to
		requiring that $\alpha^*$ satisfies the equation
		$\sum_{m\in\cM}\phi({p}^*_{t,m}) = \rho_t$.
This
is at worst a one-dimensional bisection search that embeds the previous step in each function call of the search. 
		\item Define the index set $\cN=\{m \in \cM(t)\,\,\mid\,\lambda^* \le z_m - \alpha^* \phi'(0)\}$, with $\cN=\emptyset$ if $\phi'(s)\tndni$ as $s\tndo+$. Set
		\begin{equation}\label{p_star}
		p^*_{t,m} = \left\{ 
		\begin{array}{ll}
		\frac{1}{M(t)} (\phi')^{-1}\left(\frac{z_m - \lambda^*}{\alpha^*}\right), 
		& \,\,\, m\in\cN \\
		0 & \,\,\,m\notin\cN
		\end{array} 
		\right. .
		\end{equation}
		\textbf{Return} ${P}_t^*$.
	\end{enumerate}
\end{enumerate}

\subsection{\ProofOf{\refprop{zerobias}}}
Recall that $\hat{R}_M(\theta)$ is the robust loss estimate constructed by subsampling an $M$-sized subset of the training set.
%
%
We then obtain
\begin{align*}
\E_{\tau} \left[ G(\theta)  \right] &=
\sum_{k=1}^K q_k \cdot \E \left(\frac {\Delta_k(\theta)}{q_k} + \nabla_{\theta}\hat{R}_1(\theta) \right) \\
&\hspace*{-0.35in} = \E \nabla_{\theta}\hat{R}_1(\theta) + \sum_{k=1}^K \left(\E \nabla_{\theta} \hat{R}_{2^{\tau}}(\theta) - \E \nabla_{\theta} \hat{R}_{2^{\tau-1}}(\theta)\right)  \\
&\hspace*{-0.35in} =\nabla_{\theta}R(\theta),
\end{align*}
where the telescoping sums in the last equality cancel out to leave only the leading term for $k=K$, with $\hat{R}_{2^K}(\theta) = \hat{R}_N(\theta) = R(\theta)$ by assumption.

\subsection{\ProofOf{\refthm{smalleffort}}}
The computation time in calculating $\Delta_k$ for a fixed $k$ lies mainly with estimating within $\epsilon$-accuracy the solutions to the four inner maximization problems over subsets of size up to $M(k)=2^k$.
Recall
from~\citet[Proposition 2]{gsw19}
that the computation time in obtaining approximations to a problem of size $M$ is $O(M\log M)$.
Hence, we have
\begin{align*}
\E_{\tau} T & \le C' \sum_{k=1}^K q_k \cdot  2^k k \log 2 \\
& = \frac {C' (1-r) \log 2}{1-r^{K}} \,\,  \sum_{k=1}^K k (2r)^k \\
& \le 2{C' (1-r) \log 2} \,\, \sum_{k=1}^{\infty} k (2r)^k \\
& = 2{C' r(1-r) \log 2} \,\, \sum_{k=1}^{\infty} \frac {d}{dr} ((2r)^{k}) \\
& = 2{C' r(1-r) \log 2} \,\, \frac {d}{dr} \left(\sum_{k=1}^{\infty}  (2r)^{k}\right) \\
& = 2{C' r(1-r) \log 2} \,\, \frac {d}{dr} \left(\frac 1 {1-2r}\right) \\
& = \frac {4C' r(1-r) \log 2} {(1-2r)^2}. 
\end{align*}
Here, the first inequality uses the
above $O(M\log M)$
computational complexity result.
The second inequality holds because $K>1$ and $r<1/2$ by assumption. The interchange of the derivative and the infinite sum is justified by the convergence of the sum (because $2r<1$). 

\subsection{\ProofOf{\refthm{smallvar}}}

We first provide a couple of results that will be helpful in our proof of \refthm{smallvar}, initially
addressing the feasibility of the restriction
$\tilde{P}^{*}$ of the (unique, by assumption) optimal solution $P^*$ of the
full-data problem onto the (randomly sampled) subset $\cM$, where
\begin{equation}\label{restrictPStar} 
\tilde{p}^*_m = \frac {p^*_m}{\sum_{j\in\cM}p^*_j},\,\,\,\forall m\in\cM.
\end{equation}
Denote by $\cP_{M}$ the feasibility set of \eqref{restrob} for the sampled $\cM$.
%
\begin{lemma}[{\citet[Lemma 8]{gsw19}}]\label{lem:phifeas}
	Suppose the $\phi$-divergence
	function has strictly convex level sets and
	$\rho < \bar{\rho}(N,\phi) = (1-\frac 1 N) \,\phi(\frac  N {N-1}) + \frac 1 N \phi(0)$.
	Let the $D_{\phi}$-constraint
	target $\rho_M$ of the restricted problem~\eqref{restrob} be set as stated
	in~\refthm{thm:bias}.  Then, for $M \ge M'$, we have
	\begin{equation}\label{eq:phifeas}\Prb_{\cM} (\tilde{P}^* \in \cP_{M}) \ge
	\max\left\{0, 1- \epsilon_1\right\}\cdot
	\max\left\{0, 1- \epsilon_2\right\},
	\end{equation}
	where
	\begin{align*}
	\epsilon_1 &\DefAs \eta_M^{2\delta/(1-\delta)} \sigma^2(P^*), \\
	\epsilon_2 &\DefAs \eta_M^{2\delta / (1-\delta)}\frac{\sigma^2(\phi)} {{c^2}'} , \\
	\sigma^2 (P^*) &\DefAs \frac{1}{N-1} \sum_{n=1}^N (Np^*_n-1)^2  , \\
	\sigma^2(\phi) &\DefAs \frac 1 {N-1} \sum_{n=1}^N (\phi(Np^*_n)-\rho)^2 , \\
	M' &\DefAs \min \left\{\,M\,|\, \eta_M \le \max\left\{ 1- \frac 1 {k_0} ,\,\, \zeta_0 ,\,\, \frac 1 {k_0\kappa_1}\right\}\right\} ,
	\end{align*}
	and the constants $c$ and $k_0$ are chosen such that $ c' =  (c-k_0(\kappa_1\rho+\kappa_2))/2 > 0$.
\end{lemma}
Note here that $\sigma^2(\phi)$ calculates the variance in the vector $\phi(Np^*_n)$ for \emph{any} $\phi$,
though the formula makes it resemble the $\chi^2$-divergence $\phi(s)= (s-1)^2$.

\reflem{lem:phifeas} shows that the specific choice of $\rho_M$ leads to
the restriction $\tilde{P}^*_{M}$ of the unique optimal $P^*$ to be feasible for
\eqref{restrob} with probability converging to $1$ as $M\rightarrow N$.
The following results show
that the bias in the estimation of the optimal objective is
$O_p(\eta_M)$.
\begin{lemma}[{\citet[Lemma 9]{gsw19}}]\label{lem:objbias}
	Under the assumptions of~\reflem{lem:phifeas}, there exists a $c_1>0$ such that 
	\[
	\Prb_{\cM}\left( {\eta_M}^{-1}|\hat{R}(\theta) - R(\theta)| \le c_1\right) \ge \bar{\varepsilon}
	\]
	where $\bar{\varepsilon}\DefAs\max\{0,1- \epsilon_3\} \cdot \bar{\epsilon}$, ~$\epsilon_3 \DefAs \eta_M^{2\delta / (1-\delta)}\frac{\sigma^2(R)} {{c_1^2/4}}$,
	$\sigma^2(R) \DefAs \frac 1 {N-1} \sum_{n=1}^N ((z_nNp^*_n)-\mu(R))^2 $,
	$\mu(R) \DefAs \frac 1 N \sum_n z_nNp_n$, and $\bar{\epsilon}$ is the probability on the right hand side of~\eqref{eq:phifeas}.
\end{lemma}

We will show below that $\E \left[ \| \Delta_k \|^2_2\right] \le C\,\, (2^{-k}- 2^{-K})^2$ for a fixed $k$ and a constant $C<\infty$.  
Noting that $\|a+b\|^2 \le 2(\|a\|^2 + \|b\|^2)$ , we then have
\begin{align*}
\E_{\tau} \|G(\theta)\|^2_2 & = \sum_{k=1}^K q_k \,\,\,\,\E \left\| \frac {\Delta_k}{q_k} + \nabla_{\theta}\hat{R}_1(\theta)\right\|^2_2  \\
& \hspace*{-0.3in} \le 2  \E \|\nabla_{\theta}\hat{R}_1(\theta) \|^2_2 + 2 \sum_{k=1}^K \frac 1 {q_k} \E\left\|  {\Delta_k} \right\|^2_2 \\
& \hspace*{-0.3in} \le 2  \E \|\nabla_{\theta}\hat{R}_1(\theta) \|^2_2 + \frac {2(1-r)}{1-r^K} \sum_{k=1}^K r^{-k} \left(\frac 1 {2^{k}} - \frac 1 {2^K}\right)^2 \\
& \hspace*{-0.3in} \le 2  \E \|\nabla_{\theta}\hat{R}_1(\theta) \|^2_2 + \frac {2(1-r)}{1-r^K} \sum_{k=1}^K r^{-k} 2^{-2k} \\
& \hspace*{-0.3in} \le  2 \E \|\nabla_{\theta}\hat{R}_1(\theta) \|^2_2 +  \frac {4(1-r)}{1-1/(4r)} = C(r).
\end{align*}
The final inequality follows from the assumption that $K>1$ and $r \in (1/4, 1/2)$,
whereas the first term is bounded
as in the supposition of the theorem.

Our algorithm includes the means to control the manner in which the gradient grows.
One way to ensure this is with the following additional assumption.
\begin{assumption}\label{asm:lisLsmooth}
For each $n$, the loss function $l(\cdot, \xi_n)$ 
(i) is $L$-Lipschitz smooth, i.e., has gradients $\nabla_{\theta}l(\cdot, \xi_n)$ that are $L$-Lipschitz continuous, and (ii) has a finite minima $l^*_n$ over $\theta\in\Theta$.
\end{assumption}
Here we use the common notation of $L$ for the Lipschitz constant even though it clashes with our notation for the empirical loss function;
in all cases below, the context makes the meaning of this notation clear.
Note that this assumption \emph{does not} imply the robust loss function $R(\theta)$ itself is $L$-Lipschitz smooth,
which as discussed in the main paper following the statement of \refthm{smallvar} is harder to ensure. 

Given that our proof extensively exploits the mathematical properties of statistical sampling of a finite set without replacement,
let $\{x_1,\ldots,x_N\}$ be a set of $N$ one-dimensional values with mean $\mu = \frac{1}{N} \sum_n x_n$ and variance $\sigma^2 = \frac{1}{N-1} \sum_n(x_n-\mu)^2$
and then let us sample $M<N$ of these points uniformly without replacement to construct the set $\cM= \{X_1,\ldots, X_M \} $.
We know that the probability any particular set of $M$ subsamples was chosen is ${(N-M)!\choose N!}$.
We further know that the mean $\E_{\cM}[\bar{X}] = \mu$ and the mean sample variance $\E_{\cM}[\bar{S}^2] = \sigma^2$ are both unbiased~\citep{wilks},
where $\E_{\cM}$ denotes expectation under the corresponding probability measure and $\bar{X} = \frac{1}{M} \sum_{m=1}^M X_m$ and $\bar{S}^2 = \frac{1}{M-1} \sum_{m=1}^M (X_m - \bar{X})^2$
represent the sample mean and sample variance, respectively.
Lastly, we know that the variance of the sample mean
$\E_{\cM}[(\bar{X}-\mu)^2] = (\frac{1}{M} - \frac{1}{N}) \sigma^2$
reduces to zero as $M\rightarrow N$. 

We now estimate the second moment of $\Delta_k$ for a fixed $k$,
where we start by establishing the order of $\E \| \hat{R}_{k}(\theta) - 1/2 (\hat{R}_{k,l}(\theta) + \hat{R}_{k,r}(\theta) )\|^2_2$.
%
Consider the restrictions $\tilde{P}^*_{k}$, $\tilde{P}^*_{k,l}$ and $\tilde{P}^*_{k,r}$
of the unique optimal solution $P^*$ to the sets $\cM(k)$, $\cM_l(k)$ and $\cM_r(k)$, respectively, as defined by~\eqref{restrictPStar}.
Rewrite the term 
\begin{align*}
z^t \tilde{P}^*_{k} &= \sum_{m\in\cM(k)} z_m \tilde{p}^*_{k,m} = \sum_{m\in\cM(k)}z_m\frac{p^*_{m}}{\sum_j p^*_j}
= \frac{\bar{X}(R)}{\bar{X}(P^*)}  ,
\end{align*}
where the last expression denotes the sample means obtained by sampling without replacement from 
two $N$-dimensional vectors, namely $\{x_n(R) = z_n Np^*_n\}$ and $\{x_n(P^*) = Np^*_n \}$.
We then use the Taylor expansion of the function $h(u,v) = u/v$ to obtain
\begin{align*}
&z^t \tilde{P}^*_{k} = h(\mu(R), \mu(P^*))  \\
&\quad + \nabla h(\mu(R), \mu(P^*))
{(\bar{X}(R)-\mu(R)) \choose (\bar{X}_{k}(P^*)-\mu(P^*))}\\
&\quad + {(\bar{X}_{k}(R)-\mu(R)) \choose (\bar{X}_{k}(P^*)-\mu(P^*))}^T\nabla_{\theta}^2 h(\mu(R), \mu(P^*)) \\
&\qquad\qquad \times {(\bar{X}_{k}(R)-\mu(R)) \choose (\bar{X}_{k}(P^*)-\mu(P^*))} \\
&+ o(\min\{\|\bar{X}_{k}(R)-\mu(R)\|^2,\|\bar{X}_{k}(P^*)-\mu(P^*)\|^2\}).
\end{align*}
Similar expansions apply to $z^t \tilde{P}^*_{k,l}$ and $z^t \tilde{P}^*_{k,r}$.
Noting that
\begin{align*}
\bar{X}_{k}(R) &= \frac 1 2 \left(\bar{X}_{k,l}(R)+\bar{X}_{k,r}(R)\right),\,\,\mbox{and}\\
\bar{X}_{k}(P^*) &= \frac 1 2 \left(\bar{X}_{k,l}(P^*)+\bar{X}_{k,r}(P^*)\right),
\end{align*}
we then combine the Taylor expansions to conclude
\begin{align*}
& \E \left\| z^t \tilde{P}^*_{k} - \frac 1 2 (z^t \tilde{P}^*_{k,l}+z^t \tilde{P}^*_{k,r} )\right\|^2_2 \qquad\qquad\qquad\qquad\\
&\quad = O_p{\Big(}\left(\min\left\{\E\|\bar{X}_{k}(R)-\mu(R)\|^2, \right.\right. \\
&\qquad\qquad\qquad\left.\left.\left. \,\,\E \|\bar{X}_{k}(P^*)-\mu(P^*)\,\|^2\,\,\right\}\,\,\right)^2\right)
\end{align*}
because the term containing the Hessian is the leading term that survives.

To translate this into the desired result on $\E \| \hat{R}_{k}(\theta) - 1/2 (\hat{R}_{k,l}(\theta) + \hat{R}_{k,r}(\theta) )\|^2_2$, 
we start with \reflem{lem:phifeas} which shows that the restriction $\tilde{P}^*_{k}$ is feasible for the $M(k)$ sized problem~\eqref{restrob} with high probability,
and similarly the restrictions to the two partitions are also feasible to their respective formulations.
Hence, $z^T\tilde{P}_{u}\le \hat{R}_{u}(\theta)$ for each $u=k, (k,l)$  and $(k,r)$.
Moreover, the proof of~\reflem{lem:objbias} (found in~\cite[Lemma 9]{gsw19}) argues that $\hat{R}_{u}(\theta) \le z^T\tilde{P}_{u} + c' \eta_{M(u)}$  for the same $u$.
Thus, $z^T\tilde{P}_u \le \hat{R}_u(\theta) \le z^T \tilde{P}_u + c'\eta_{M(u)}$ with high probability, for each $u$, and we therefore have
\begin{align*}
& \E \left\| \hat{R}_{k} (\theta) - \frac 1 2 \left( \hat{R}_{k,l}(\theta)+ \hat{R}_{k,r}(\theta)\right)\right\|^2_2 \\
& \qquad = O_p\left(\left(\frac 1 {2^k} - \frac 1 {2^K}\right)^2\right),
\end{align*}
where the last expression comes again from the variance relation for sampling uniformly without replacement from finite sets (of size $N=2^K$). 

The desired result for $\Delta_k$ follows by observing the form, for each $u$, of $\hat{R}_{u}(\theta) = \sum_{m\in\cM(u)} l (\theta,\xi_m) \hat{p}^*_{u,m}$ in terms of its unique optimal solution $\hat{P}^*_u$. 
Define 
$$g(\theta)\,\,\DefAs\,\, \hat{R}_{k} (\theta) - \frac 1 2 \left( \hat{R}_{k,l}(\theta)+ \hat{R}_{k,r}(\theta)\right),
$$
%
whose gradient $\Delta_k$ we seek and whose form is obtained from Proposition~\ref{prop:rgrad}.
This involves applying Danskin's theorem to obtain that, for each $u$,
$$
\nabla \hat{R}_{u}(\theta) = \sum_{m\in\cM(u)} \nabla_{\theta} l (\theta,\xi_m) \hat{p}^*_{u,m}.$$
Taking the derivative of $g(\theta)$ with respect to $\theta$, we have
\begin{align*}
\nabla_{\theta} g(\theta) = \sum_{m_1\in\cM_l(k)} \nabla_{\theta} l(\theta,\xi_{m_1}) \left(\hat{p}^*_{m_1} - \frac {\hat{p}^*_{l,m_1}} 2 \right)
\,\,\,+ \\
\qquad \sum_{m_2\in\cM_r(k)} \nabla_{\theta} l(\theta,\xi_{m_2}) \left(\hat{p}^*_{m_2} - \frac {\hat{p}^*_{r,m_2}} 2\right),
\end{align*}
where $(\hat{p}^*_m)$, $(\hat{p}^*_{l,m})$ and $(\hat{p}^*_{r,m})$ represent the optimal solutions to the approximate inner maximization over sets $\cM(k)$, $\cM_l(k)$ and $\cM_r(k)$, respectively.

A special implication of the additional Assumption~\ref{asm:lisLsmooth} that each $l(\theta,\xi_n)$ is $L-$Lipschitz smooth is that 
\[
\|\nabla_{\theta}l(\theta,\xi_n)\|^2 \le 2L \left(l(\theta) - l^*_n\right).
\]
Letting $l^{**} = \min_n l^*_n$, we therefore obtain
\begin{align*}
\|\nabla_{\theta} g(\theta)\|^2 &\le\,\, 2L \left( \sum_{m_1\in\cM_l(k)} l(\theta,\xi_{m_1}) \left(\hat{p}^*_{m_1} - \frac {\hat{p}^*_{l,m_1}} 2 \right)
\right.\\
&\qquad +\,\,\sum_{m_2\in\cM_r(k)} l(\theta,\xi_{m_2}) \left(\hat{p}^*_{m_2} - \frac {\hat{p}^*_{r,m_2}} 2\right)\\
& \qquad-\,\, l^{**} \left(\sum_{m\in\cM(k)}p^*_m - \frac 1 2 \sum_{m_1\in\cM_l(k)}\hat{p}^*_{l,m_1}\right.\\
&\qquad\quad\left.\left. - \frac 1 2 \sum_{m_2\in\cM_r(k)}\hat{p}^*_{r,m_2}\right)\right) \\
&\le 2 L g(\theta),
\end{align*}
and thus the same order of the variance of $g(\theta)$ applies to $\Delta_k$.

\section{Numerical Experiments}\label{apdx:expt}

\subsection{Algorithmic Performance}
We first consider numerical experiments to empirically evaluate our GSSG in comparison with FSG and PSSG, expanding upon the empirical results presented in the main paper.
The framework and setting for these numerical experiments is exactly as described in Section~\ref{ssec:alg} while varying the value of $\rho$.
In addition, all the plots in this Appendix are over cumulative computational time, which provides a more accurate picture of the running times of the various methods. 
We focus on the two datasets considered in the main paper,
namely the HIV-1 Protease Cleavage dataset ($N\sim 10^4$ samples) from~\cite{UCI} and the Reuters Corpus Volume I dataset ($N\sim 10^6$ samples) from~\cite{RCV1}.
Each are considered in turn.

We re-implemented the methods of ~\cite{nd17} and~\cite{gsw19} since full source code was not made available with the corresponding publications,
and will open our github repository for all code used to produce these results for public consumption once the anonymity requirement is resolved.

\subsubsection{HIV-1 Protease Cleavage}
Starting with the HIV-1 protease cleavage dataset,
recall that this dataset helps develop effective protease cleavage inhibitors by predicting whether the HIV-1 protease will
cleave a protein sequence in its central position
or not.
Additional details are as described in Section~\ref{sssec:HIV}.

\begin{figure*}[htbp]
	\vskip -0.1in
	\begin{center}
		\includegraphics[width=0.325\textwidth]{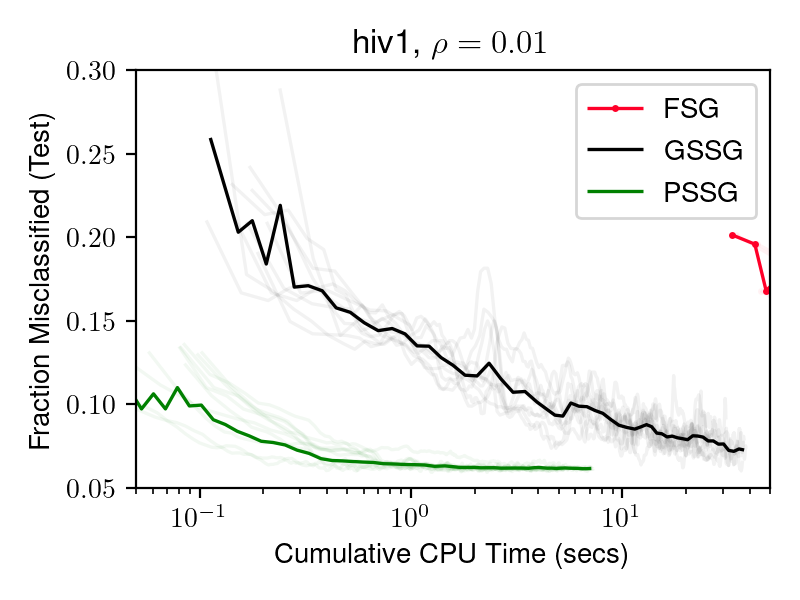}
		\includegraphics[width=0.325\textwidth]{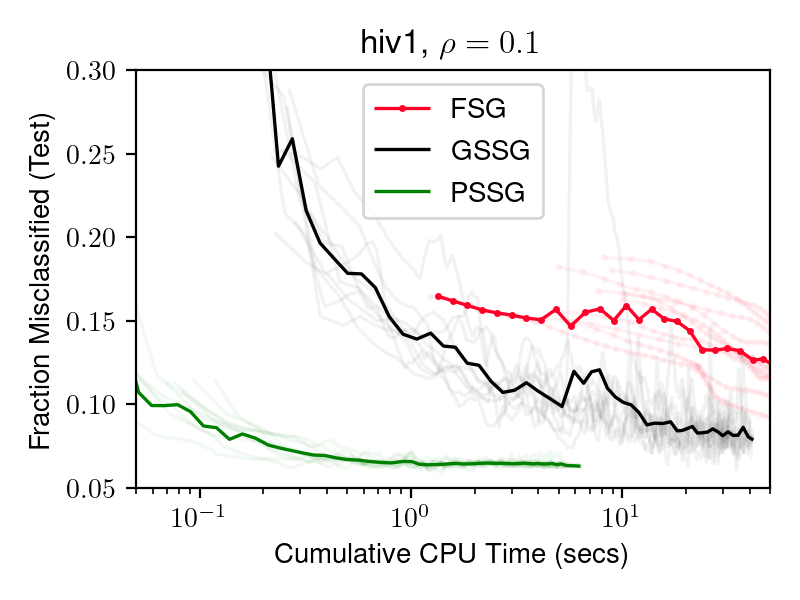}
		\includegraphics[width=0.325\textwidth]{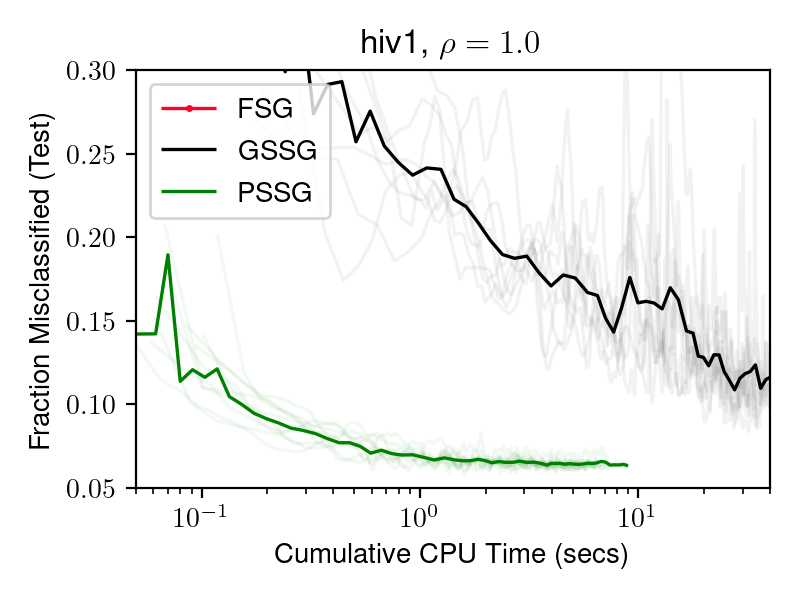}
		\caption{Comparison of GSSG (black), PSSG (green),
                  FSG (red) on misclassification in testing
                  versus cumulative computational time (CPU time in seconds) over HIV-1 dataset with $\rho=0.01$ (left), $\rho=0.1$ (center) and $\rho=1.0$ (right).}
		\label{fig:cpuHIV}
	\end{center}
	\vskip -0.2in
\end{figure*}

Figure~\ref{fig:cpuHIV} presents a comparison of the fractional misclassification
performance of DSSG, FSG and GSSG over the testing data from $10$ numerical experiments, where each run uses a
different random partition of the data into training and testing datasets.
As noted in the captions, the left panel compares the methods for a $D_{\phi}$ constraint parameter $\rho = 0.01$, while the central panel employs $\rho=0.1$ and the right $\rho=1.0$. Our proposed GSSG algorithm considerably outperforms the FSG algorithm for each value of $\rho$ considered.
As observed in the main paper, GSSG suffers from higher variance in its sample paths, and hence does not work as well as the PSSG for this small dataset.
The plots here are over the cumulative computational time in CPU seconds;
observe that each iteration of GSSG is more expensive (in CPU seconds) as compared to the iterations of PSSG, and hence the difference in the center figure when contrasted with Figure~\ref{fig:rho0.1:HIV} in the main paper.

\subsection{Reuters RCV1}
Turning to the Reuters Corpus Volume I (RCV1) dataset,
recall that the purpose of this dataset is to classify each sample article as either belonging to a corporate/industrial category or not based on its content,
representing a substantially larger and more challenging dataset than HIV.
Additional details are as described in Section~\ref{sssec:RCV}.

\begin{figure*}[htbp]
	\vskip 0.2in
	\begin{center}
		\includegraphics[width=0.325\textwidth]{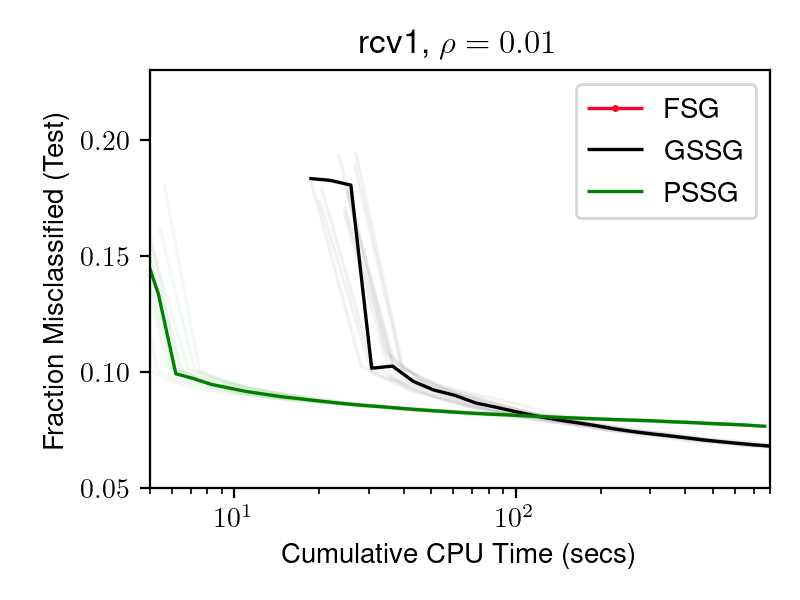}
		\includegraphics[width=0.325\textwidth]{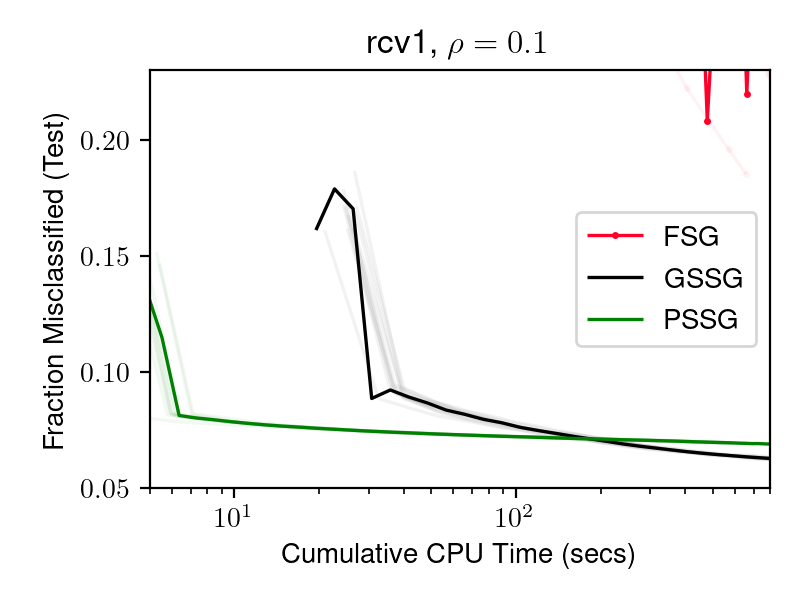}
		\includegraphics[width=0.325\textwidth]{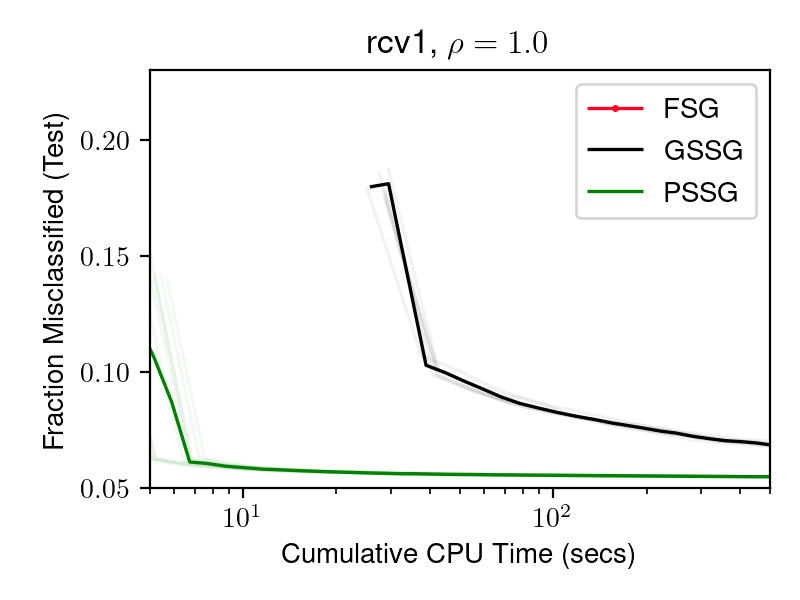}
		\caption{Comparison of GSSG (black), PSSG (green),
			FSG (red) on misclassification in testing
			versus cumulative computational time (CPU time in seconds) over HIV-1 dataset with $\rho=0.01$ (left), $\rho=0.1$ (center) and $\rho=1.0$ (right).}
		\label{fig:cpuRCV}
	\end{center}
	\vskip -0.2in
\end{figure*}

We consider here an additional set of empirical results from the $10$ numerical experiments.
Specifically, Figure~\ref{fig:cpuRCV} contrasts the performance of the GSSG algorithm with PSSG and FSG for $\rho=0.01$ (left panel) , $\rho=0.1$ (center panel) and $\rho=1.0$ (right panel).
In the left two cases, GSSG takes up to two orders of magnitude less CPU time than FSG, and is also observed to reduce the time taken to converge over PSSG.
Contrast the lower variance experienced by GSSG in each of these $\rho$ cases to that in the previous dataset HIV1. This points to the importance of the variance of the sample paths in the efficacy of GSSG over PSSG.

\subsection{Generalization Performance}
We next consider numerical experiments on model generalization that compare our GSSG with PSSG and ERM, expanding upon the empirical results presented in the main paper.
The framework and setting for these numerical experiments is exactly as described in Section~\ref{ssec:gen} while varying the value of $\rho$ and staying faithful to the broad guideline that $\rho = O(\sqrt{d/N})$.

\begin{table*}[!htbp]
	\vskip -0.1in
	\setlength{\tabcolsep}{5pt}
	\centering
	\begin{tabular}{l||r|r|r||r|r||r|r||r}
		\hline
 &	&	&	&\multicolumn{2}{c||}{$\rho = 0.1$}	&\multicolumn{2}{c||}{$\rho = 0.01$} \\ 
\cline{5-8}
Dataset	&$d$	&$N$	&$\sqrt{d/N}$	&{GSSG}	&{PSSG}		&{GSSG} &{PSSG} 	&{Reg. ERM } \\ 
\hline\hline
adult	&119	&45222	&0.05		&${17.1 \pm 0.3}$	& $\mathbf{16.6\pm 0.1}$ &$\mathbf{16.3 \pm 0.4}$	& $\mathbf{16.6\pm 0.2}$	& $\mathbf{16.7\pm 0.0}$ \\
imdb.drama	&1001	&120919	&0.09 	&${37.1 \pm 0.7}$	&$\mathbf{36.2\pm 0.1}$  	&$\mathbf{36.1 \pm 0.8}$  &$\mathbf{36.2\pm 0.1}$  &$37.1\pm 0.0$\\ 
hiv1	&160	&5830	&0.17		&$\mathbf{6.1 \pm 0.5}$ & $5.9\pm 0.1$	&$\mathbf{5.7 \pm 0.4}$ & $5.8\pm 0.0$	& $\mathbf{5.6\pm 0.0}$ \\
rcv1	&47236	&804414 &0.24	 	&$\mathbf{5.5 \pm 0.1}$	 &$\mathbf{5.4\pm 0.0}$ &$\mathbf{5.5 \pm 0.2}$	 &${7.1\pm 0.0}$  &$5.6\pm 0.0$ \\
fabert	&800	&8237	&0.31	&$\mathbf{10.4 \pm 0.4}$	& $\mathbf{9.8\pm 0.0}$	&$\mathbf{10.4 \pm 0.5}$	& $\mathbf{10.0\pm 0.2}$	& $10.1\pm 0.0$ \\ 
dilbert	&2000	&10000	&0.45	 	&$\mathbf{1.6 \pm 0.6}$ &$\mathbf{1.2\pm 0.1}$  &$\mathbf{1.5 \pm 0.5}$ &$\mathbf{1.2\pm 0.1}$  &$\mathbf{1.2\pm 0.0}$ \\ 
guillermo	&4296	&20000	&0.46		&${32.6 \pm 1.5}$  &$\mathbf{30.2\pm 0.5}$ &${33.4 \pm 1.8}$ &$31.1\pm 0.1$  &$\mathbf{30.7\pm 0.1}$ \\ 
riccardo	&4296	&20000	&0.46		&${2.4 \pm 0.3}$ &$\mathbf{1.6\pm 0.0}$ &${2.0 \pm 0.4}$ &${2.1\pm 0.1}$  &$\mathbf{1.5\pm 0.0}$  \\ 
gina\_prior	&784	&3468	&0.48		&${14.4 \pm 0.8}$ &$ 13.0\pm 0.5$	&${13.6 \pm 0.8}$ &$ 12.7\pm 0.5$	& $\mathbf{11.8\pm 0.1}$ \\
gina\_agnostic	&970	&3468	&0.53	 &${14.8 \pm 1.8}$	&$13.9\pm 0.3$ &${14.1 \pm 1.2}$  &$13.2\pm 0.3$  &$\mathbf{12.6\pm 0.1}$ \\ 
Bioresponse	&1776	&3751	&0.69	  &${25.6 \pm 1.4}$ &$24.2\pm 0.4$ &${24.5 \pm 0.9}$ &$23.7\pm 0.6$  &$\mathbf{21.6\pm 0.2}$\\ 
la1s.wc	&13195	&3204	&2.01	  	&${10.3 \pm 1.4}$ &$\mathbf{8.3\pm 0.2}$ &$\mathbf{9.1 \pm 0.9}$ &${8.9\pm 0.1}$  &$8.5\pm 0.0$ \\
OVA\_Breast	&10935	&1545	&2.73	&${4.1 \pm 0.6}$ &$3.0\pm 0.1$ &${3.6 \pm 0.6}$  &$2.9\pm 0.1$  &$\mathbf{1.8\pm 0.1}$ \\
tr31.wc	&10128	&927	&3.32		&$\mathbf{2.9 \pm 0.5}$ &$\mathbf{2.6\pm 0.3}$ &$\mathbf{2.8 \pm 0.4}$ &$\mathbf{2.8\pm 0.3}$  &$\mathbf{2.7\pm 0.1}$ \\ 
\hline
	\end{tabular}
\caption{Comparison of the two DRL algorithms (GSSG and PSSG) and regularized ERM formulations over $14$ publicly available machine learning (ML) datasets,
from UCI$^{\ast}$ \cite{UCI}, OpenML$^{\dag}$ \cite{openml} and SKLearn$^{\ddag}$ \cite{RCV1},
arranged in increasing value of the third column. 
The first pair of columns provides the $d$ and $N$ characteristics of each dataset. The third column provides $\sqrt{d/N}$, the characteristic that~\citeN{bwm16,nd17} provide as a guideline for setting $\rho$ in~(\ref{absfmln}). The fourth and fifth columns provide a 95\% confidence interval (CI) of the percentage misclassified over withheld test datasets for $\rho=0.1 $ as estimated by the GSSG and PSSG methods respectively. The sixth and seventh columns provide the same results for $\rho=0.01$,  while the final column provides the same for the $10$-fold CV regularized ERM method. The best-performing method (within its CI) is highlighted in bold. Note that the fifth, seventh and final columns are repeated from Table~\ref{table:gen}.}
\label{table:apdxgen}
\vskip -0.2in
\end{table*}

Table~\ref{table:apdxgen} presents the $d$, $N$ and $\sqrt{d/N}$ values for each dataset.
Accordingly, we reran the two DRL algorithms GSSG and PSSG with
$\rho=0.01$ and
$\rho=1.0$
for all the datasets.
However, based on the results in the previous section, we focus here on the results for the values of $\rho=0.1$ and $\rho=0.01$
and we present the corresponding latter results in columns six and seven of Table~\ref{table:apdxgen}. 
The actual values in the third column suggest that, in several datasets, a value of $\rho=0.01$ might possibly be preferable.

Note first that the $\rho=0.01$ results of the DRL methods does help match the generalization performance of the regularized ERM method or further improve the DRL performance in a handful of cases.
The GSSG method, in this set of results, also seems to improve on  the performance of PSSG in many cases (adult, hiv1, rcv1, dilbert, riccardo, gina\_prior, gina\_agnostic, Bioresponse, la1s.wc, OVA\_Breast, tr31.wc).
For the cases of gina, gina\_prior and Bioresponse, ERM continues to provide better performance, but we note here that the best obtained test misclassification percentage is high,
indicating that the efficacy of linear models with the logistic loss function may be limited in these cases.
Each run of DSSG for $\rho=0.01$ took the same order of computational effort as the runs for $\rho=0.1$.
Hence, the added benefit of a few extra DRL runs is obtained without significantly multiplying the computational time.
In practice, since the constants associated with the $\rho = O(\sqrt{d/N})$ guideline are hard to compute,
one may wish to try a couple of $\rho$ values relatively close by to determine if additional improvements in performance can be obtained.

\end{document}